\documentclass[lettersize,journal]{IEEEtran}
\usepackage{amsmath,amsfonts}
\usepackage{algorithmic}
\usepackage{algorithm}
\usepackage{array}
\usepackage{textcomp}
\usepackage{stfloats}
\usepackage{url}
\usepackage{verbatim}
\usepackage{graphicx}
\usepackage{cite}
\usepackage{ragged2e}
\usepackage{enumerate}
\usepackage{multirow}
\usepackage{color}
\usepackage{booktabs}
\usepackage{subfigure}
\usepackage{colortbl}
\usepackage{xcolor}
\begin{document}

\title{DFREC: DeepFake Identity Recovery Based on Identity-aware Masked Autoencoder}

\author{Peipeng Yu,~Hui Gao,~Jianwei Fei,~Zhitao Huang,~Zhihua Xia,~\IEEEmembership{Member,~IEEE, }Chip-Hong Chang, \IEEEmembership{Fellow,~IEEE}
        % <-this % stops a space
\thanks{Manuscript received March 6, 2025. This work is supported in part by the National Key R\&D Program of China under Grant number 2022YFB3103100, in part by the National Natural Science Foundation of China under grant numbers 62122032 and U23B2023, in part by the Jinan University Outstanding and Innovative Doctoral Student Training Project under Grant number 2023CXB021, and in part by the Ministry of Education, Singapore, under its Academic Research Fund (AcRF) Tier 2 under Award MOE-T2EP50220-0003.}        
\thanks{Peipeng Yu, Hui Gao, Zhitao Huang, and Zhihua Xia are with College of Cyber Security, Jinan University, Guangzhou, 510632, China. Jianwei Fei is with Department of Computer and Information Science, University of Macau, Macau, China. Chip-Hong Chang is with the School of Electrical and Electronic Engineering, Nanyang Technological University, Singapore 639798.}% <-this % stops a space
\thanks{Corresponding authors: Zhihua Xia and Chip-Hong Chang. (e-mail: xia\_zhihua@163.com, echchang@ntu.edu.sg)}}

\markboth{Journal of \LaTeX\ Class Files,~Vol.~14, No.~8, August~2021}%
{Shell \MakeLowercase{\textit{et al.}}: A Sample Article Using IEEEtran.cls for IEEE Journals}


\maketitle

\begin{abstract}
  Recent advances in deepfake forensics have primarily focused on improving the classification accuracy and generalization performance. Despite enormous progress in detection accuracy across a wide variety of forgery algorithms, existing algorithms  lack intuitive interpretability and identity traceability to help with forensic investigation. In this paper, we introduce a novel DeepFake Identity Recovery scheme (DFREC) to fill this gap. DFREC aims to recover the pair of source and target faces from a deepfake image to facilitate deepfake identity tracing and reduce the risk of deepfake attack. It comprises three key components: an Identity Segmentation Module (ISM), a Source Identity Reconstruction Module (SIRM), and a Target Identity Reconstruction Module (TIRM). The ISM segments the input face into distinct source and target face information, and the SIRM reconstructs the source face and extracts latent target identity features with the segmented source information. The background context and latent target identity features are synergetically fused by a Masked Autoencoder in the TIRM to reconstruct the target face. We evaluate DFREC on six different high-fidelity face-swapping attacks on FaceForensics++, CelebaMegaFS and FFHQ-E4S datasets, which demonstrate its superior recovery performance over state-of-the-art deepfake recovery algorithms. In addition, DFREC is the only scheme that can recover both pristine source and target faces directly from the forgery image with high fadelity.
\end{abstract}


\section{Introduction}
\label{sec:intro}
\IEEEPARstart{T}{he} emergence of deepfake technology, particularly its face-swapping capabilities, has ushered in an unprecedented era of digital media manipulation. Since its inception, deepfake has rapidly evolved into tools for disseminating misinformation, identity theft, and fraud. The growing ethical and legal concerns on the widespread dissemination of deepfake content have underscored the urgency of developing robust deepfake forensic tools. This is crucial not only for safeguarding individuals' rights but also upholding the information integrity in the digital age.

\begin{figure}[t]
	% \centering
	\subfigure[Deepfake Face Swapping Model Training]{
		\begin{minipage}[b]{0.48\textwidth}
		\includegraphics[width=\textwidth]{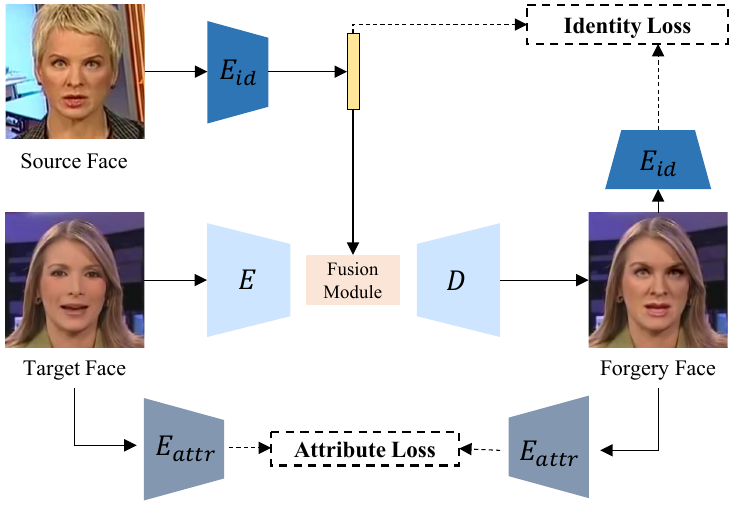}
		\end{minipage}
		}
	\subfigure[Identity similarity between forgery and target/source faces]{
		\begin{minipage}[b]{0.48\textwidth}
			\includegraphics[width=\textwidth]{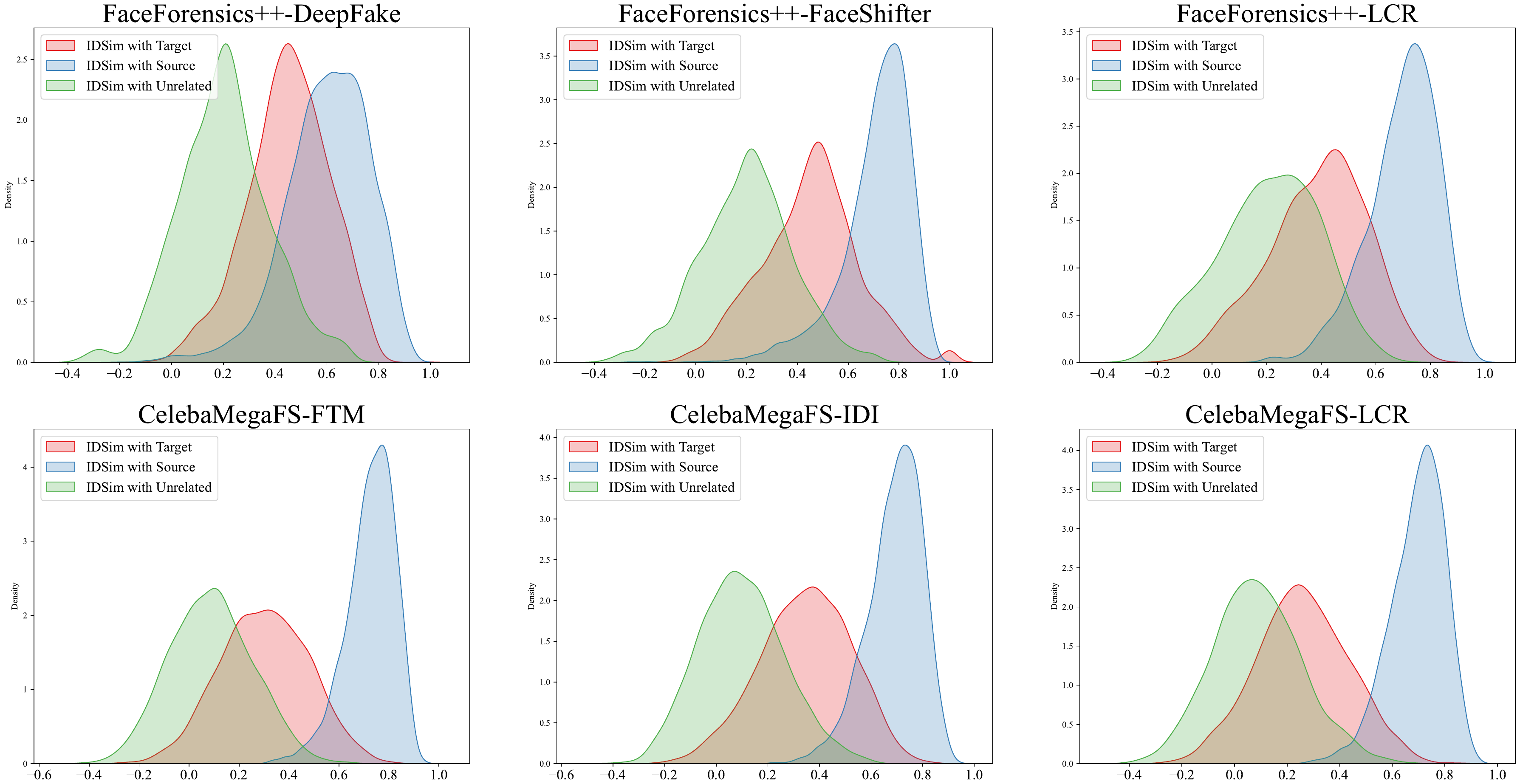}
		\end{minipage}
		}
	\caption{(a) The training process of deepfake face swapping models. The identity loss is used to increase the similarity between the forgery and source face identities, and the attribute loss is used to ensure the similarity between the forgery and target face attributes.  (b) The probability density functions of the identity similarity (IDSim) between the forgery and source, forgery and target, and forgery and other unrelated faces. The results are calculated on the five types of forgeries on the FaceForensics++ and CelebaMegaFS datasets.
		}
		\label{fig:distribution}
\end{figure}

Most of the existing deepfake forensic methods focus on improving the generalization detection performance across various kinds of forgery images \cite{shiohara2022detecting,dong2022protecting,liu2023ti2net,zhao2023istvt,lin2024preserving,ba2024exposing,10364845}. While these methods demonstrate impressive performance in classification and localization tasks, they are inherently limited in their ability to provide comprehensive forensic evidence. Specifically, they fail to establish a coherent chain of evidence linking the detection results to the identities involved in the manipulation. In other words, they cannot reliably trace the source identity (i.e., the person whose facial features are transferred) and the target identity (i.e., the individual whose face appears in the manipulated content). The inability to trace the origin of deepfake content can delay verifying the origin of deepfake content and mitigating potential damage. 
More importantly, as the onus of proof is borne by the victims with source identities, an accurate recovery of the source and target identities could help to collect and identify evidences for pursuing legal action against the defendant responsible for creating and disseminating deepfake content with malicious intent.

Motivated by this forensic investigation gap, we propose to approach the deepfake forensic task from the perspective of pristine face recovery. While deepfake algorithms have evolved through various iterations, their core operation involves transferring the source identity onto the target face while retaining the attributes of the target face (such as expression, pose, lighting, etc.). This synthesis process blends the identities of both faces, resulting in a composite image that incorporates features from both identities \cite{huang2023implicit}. As illustrated in Fig. \ref{fig:distribution}, the identity similarity between the forgery face and the target face is closely resembling the similarity between the forgery face and the source face. This observation suggests that the forgery face encompasses a portion of the target face information. Building on this insight, we propose DFREC, a DeepFake Identity Recovery scheme based on Identity-aware Masked autoencoder, which is designed to recover both the source and target faces. The main contributions of this paper are outlined as follows:

\begin{itemize}

	\item We propose a digital forensic aid DFREC to restore the identities of both target and source faces. DFREC accomplishes this task through ($i$) designing an Identity Segmentation Module (ISM) to segment information related to the source and target faces respectively, ($ii$) constructing a Source Identity Recovery Module (SIRM) to recover the source face using the segmented source information, and ($iii$) designing a Target Identity Recovery Module (TIRM) to recover the target face using the background information and latent target identity features extracted by the SIRM. Through identity segmentation and recovery, DFREC is able to restore the pristine source and target faces with high fidelity for intuitive interpretation.

	\item We propose an Identity-aware Masked Autoencoder (MAE) to integrate latent target identity features into the face recovery process. Specifically, the MAE utilizes segmentation maps generated by the ISM to mask the input image and uses the Vision Transformer (ViT) for feature extraction. The target identity features are incorporated into the visible patch embedding to guide the MAE to restore pristine quality image with the desired target face identity.

	\item We evaluate DFREC on the FaceForensics++, CelebaMegaFS, and FFHQ-E4S datasets, across six types of deepfake face swapping approaches. Compared to existing face inpainting methods and deepfake recovery methods, our approach achieves superior face recovery performance. By successfully recover both the pristine source and target faces with the correct identities, DEREC provides a non-repudiable intrinsic evidence of provenience that is gleaned directly from the forgery image.
\end{itemize}

\section{Related Works}
In this section, we first review existing deepfake face-swapping algorithms, then discuss current detection methods, and finally introduce the development of deepfake recovery techniques.
\subsection{Deepfake Face Swapping}

Face swapping is a representative algorithm in deepfake technology, which leverages generative models to encode target features and synthesize the swapped face. Convolutional neural networks (CNNs) and generative adversarial networks (GANs) are often employed to achieve identity-attribute encoding~\cite{gu2019mask,li2020advancing,chen2020simswap,zhu2021megafs}. Gu \emph{et al.} \cite{gu2019mask} proposed to leverage conditional GANs to generate specific swapping faces. This framework learns feature embeddings for each facial component independently, enabling precise and customizable local face editing. To further improve the visual quality of the swapped faces, Li \emph{et al.} proposed FaceShifter \cite{li2020advancing}, which employs a U-Net architecture to capture the target face’s attributes. It incorporates an Adaptive Embedding Integration Network to blend source facial features with target attributes, producing photorealistic face-swapping results. Similarly, SimSwap~\cite{chen2020simswap} introduces a Weak Feature Matching Loss to optimize generated faces. It exploits both shallow and deep discriminator features for better consistency and fidelity. Based on these advancements, Zhu \emph{et al.}~\cite{zhu2021megafs} proposed MegaFS, a novel approach that operates in an extended latent space to generate megapixel-level swapped images. It achieves superior identity swapping while preserving high-resolution facial attributes. Recent advances \cite{xu2022region,liu2023fine,han2025face} further enhance realism and preserve the geometry and texture priors, as well as other non-facial attributes in swapped images.

It is worth noting that these advanced face swapping algorithms produce high-fidelity forgery face identity by preserving the target's face attributes (such as expression and pose). Due to the inherent difficulty in decoupling identity features from attribute features, the extracted attribute features often exhibit a certain degree of correlation with identity~\cite{huang2023implicit}. This inherent correlation often leads to a fusion of source and target identities. Inspired by this insight, our proposed DFREC scheme extracts latent identity features from forgery images to recover the source and target faces, thereby enabling the tracing of both source and target identities.

\begin{figure*}[tbp]
	\centering
	\includegraphics[width=\linewidth]{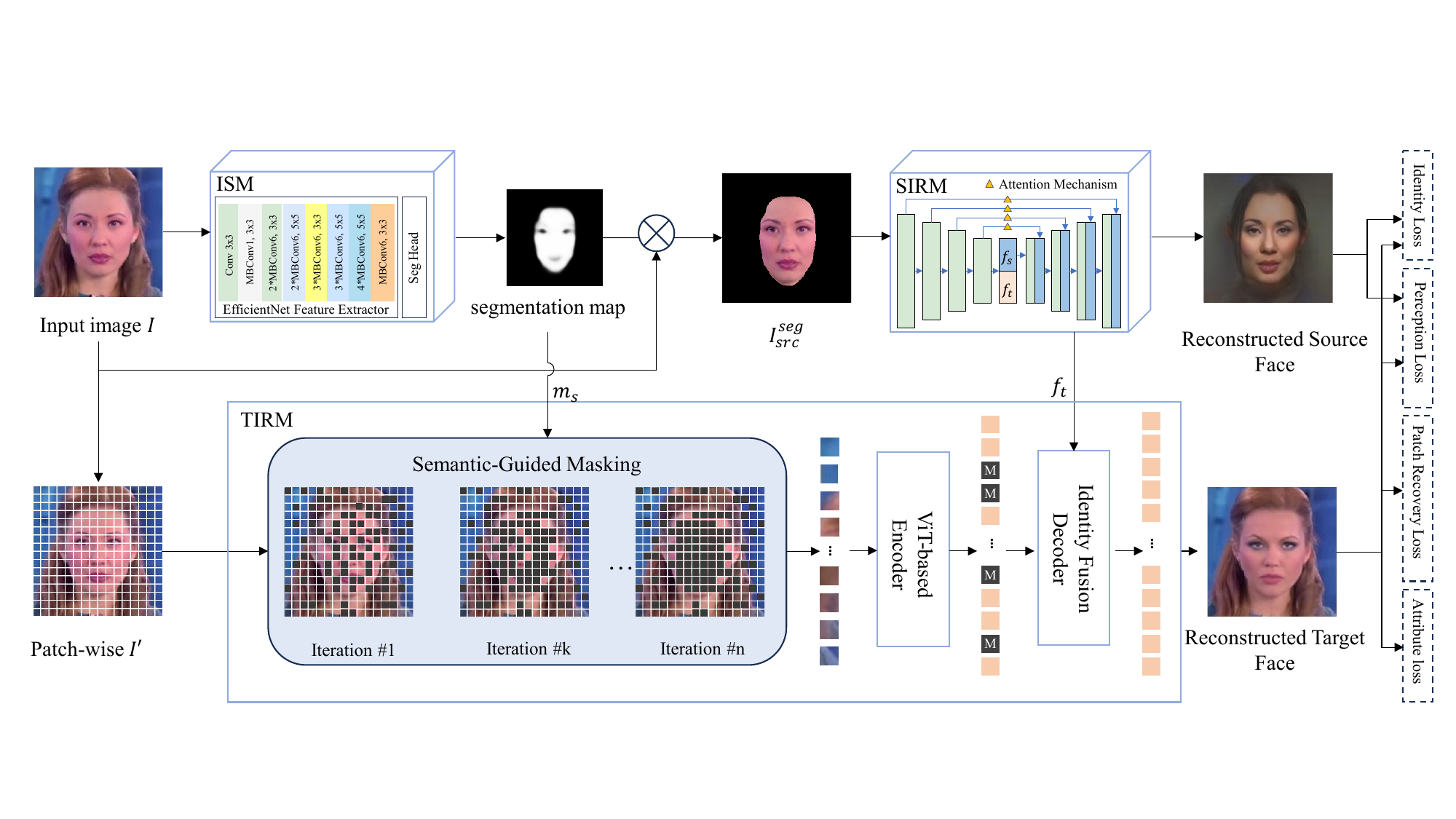}
	\caption{An overview of DFREC. The Identity Segmentation Module (ISM) segments an input image to extract the source and target information. The Source Identity Recovery Module (SIRM) and Target Identity Recovery Module (TIRM) recover the source and target identities, respectively.
	}
	\label{fig:scheme}
\end{figure*}

\subsection{Deepfake Forgery Detection}

Deepfake forgery detection predominantly utilizes CNN-based methodologies, and focus more on the generalization performance in recent years \cite{10458687}. Generalization through diverse strategies, including data augmentation~\cite{li2020face,shiohara2022detecting,nguyen2024laa} and feature engineering~\cite{zhao2021learning, ni2022core, yan2024transcending, 10054130, 10286083}, have been explored to detect fine-grained photorealistic face forgery. In the realm of data augmentation, Li \emph{et al.} \cite{li2020face} proposed a method that selects two facial images based on the similarity of their landmarks, generating blended faces by applying randomly created masks on facial regions. This strategy significantly improved the generalization performance of detection models. Shiohara \emph{et al.} \cite{shiohara2022detecting} further introduced simulation based inference (SBI) to simulate challenging forgery images by transforming  local regions of pristine images. Building on this foundation, Nguyen \emph{et al.} \cite{nguyen2024laa} proposed LAANet, which calculates vulnerable points in the SBI framework and employs an attention mechanism to enhance sensitivity to these regions, achieving state-of-the-art generalization accuracy.
In terms of feature engineering, Zhao \emph{et al.} \cite{zhao2021learning} enhanced forgery detection by constraining the correlation between features of authentic and forged regions to improve the model's perception of manipulated content. Ni \emph{et al.} \cite{ni2022core} developed a method that enforces feature consistency across diverse augmentation operations, which facilitates the learning of stable and reliable representations. More recently, Yan \emph{et al.} \cite{yan2024transcending} proposed to expand the diversity of feature representations by transforming latent space, which effectively improves the detection performance against previously unseen forgeries.
Despite their high detection accuracy, these forensic tools cannot trace the origin of deepfake content, which hinders the victims from pursuing legal action and limits their applicability in judicial contexts.

\subsection{Deepfake Recovery}

Different from detection-based methods, deepfake recovery methods aim to reconstruct the original target face. Existing studies primarily leverage face reconstruction to facilitate deepfake detection.  To improve the identity-relevant deepfake detection and interpretability, RECCE \cite{cao2022end} proposed a reconstruction learning approach to discern compact real data representations. They trained an autoencoder with real face images to reconstruct faces and then utilized the face reconstruction error to guide the training of deepfake models. Delocate~\cite{hu2024delocate}, on the other hand, randomly occluded local regions of faces and employed a Masked Autoencoder to reconstruct the original facial components, thereby assisting  deepfake detection. Additionally, DFI~\cite{ai2023deepreversion,ai2023deepfake} proposed to trace the original target face by decoupling identity and attributes. However, RECCE and Delocate are limited to generating distorted facial images and cannot effectively recover the target identity from the forgery face. DFI cannot avoid the disturbance of source identities on the recovered target faces. To tackle these limitations, we first segment the two types of information within forgery faces to mitigate their interference, and then integrate the background information with the implicit target identity features to recover the deepfake identity.

\section{The Proposed Method}

\subsection{Overview}
Deepfake face swapping images contain both the identity of the source face and some attributes of the target face (expression, pose, lighting, etc.). Given the existence of correlation between attributes and identity, we believe that the source and target identities of a face-swapped image can be segregated to allow for the reconstruction of both the source and target faces. 

We name our proposed deepfake identity recovery scheme DFREC. As illustrated in Fig. \ref{fig:scheme}, DFREC consists of three components: an Identity Segmentation Module (ISM), a Source Identity Reconstruction Module (SIRM), and a Target Identity Reconstruction Module (TIRM). The ISM classifies each pixel of an input image according to its probability of affiliation with the source face. By taking the dot product between the input image and the output probability map, the segmented source face information is produced. The SIRM reconstructs the source face and extracts the target identity features from the decomposed source face information simultaneously. The TIRM masks the image using the segmentation map and recovers the target face by combining the background information with the target identity features using the Masked Autoencoder (MAE). A quadruple loss that combines Identity Loss, Perceptual Loss, Attribute Loss and Patch Recovery Loss is used to guide the face recovery to achieve state-of-the-art (SOTA) recovery quality of the source and target faces.

\subsection{Identity Segmentation Module (ISM)}
The ISM is designed to segment source and target face information embedded in a face-swapping image. Specifically, we employ the DeepLabv3 architecture based on atrous convolution, which consists of a feature extractor, a segmentation head and a classification head. The classification head determines if an input facial image is a pristine or a forgery face while the segmentation head produces a probabilistic map for the extraction of source face information. We use pretrained EfficientNet-B0 as in~\cite{nirkin2021hyperseg} to extract both low-level and high-level features. The high-level features are then directed into the classification head for forgery classification. Meantime, the high-level features are supplied to the segmentation head for up-sampling and combined with the low-level features to generate the segmentation map. As illustrated in Fig. \ref{fig:scheme}, it takes the image $I$ as input and generates a segmentation map $m_s$. This map is used to produce a segmented image $I_{src}^{seg}$ for source face recovery as well as guide the MAE to iteratively mask source-related patches for the target face recovery. It is worth noting that we refrain from utilizing any loss function to constrain the segmentation process, as it will autonomously compute the necessary information driven by subsequent tasks.

\subsection{Source Identity Recovery Module (SIRM)}

The SIRM is designed to recover the source face and extract the latent target identity features using the segmented image $I_{src}^{seg}$. Specifically, we construct a UNet-like model for source face recovery. As depicted in Fig. \ref{fig:scheme}, the SIRM is made up of an encoder and a decoder. The encoder is responsible for identity feature extraction. The extracted identity features are then utilized by the decoder for the source face recovery. Due to the fusion of two identities, features extracted by the encoder may inevitably contain some target face features. This may detrimentally mislead the source face recovery. To address this problem, the features extracted by the encoder are divided into two parts: source identity feature $f_{s}$ and target identity feature $f_{t}$.  The separated $f_{s}$ is independently exploited by the decoder for the recovery of the source face, and the separated $f_t$ is utilized for subsequent target face recovery.
An attention mechanism (represented by $\triangle$ in Fig. \ref{fig:scheme}) is then introduced into the skip connection to preserve features pertinent to the source face. Finally, the source face is reconstructed from the segmented source information. 

% To enhance the realism of the recovered source face, the UNet-like model is trained using the following two loss functions.

\subsection{Target Identity Recovery Module (TIRM)}
\label{sec:target}

The TIRM is devised to recover the target face image. Given that deepfake face swapping algorithms aim to eliminate as much of the target identity from the swapped image as possible, restoring the target face poses more challenges than restoring the source face. To tackle this issue, we introduce an Identity-aware Masked Autoencoder (MAE) to combine the background information and the predicted target identity features to recover the target face. As illustrated in Fig. \ref{fig:scheme}, the TIRM comprises two stages: Semantic-Guided Masking and Identity Recovery.

\subsubsection{Semantic-Guided Masking}
This stage masks a fixed proportion of patches under the guidance of the predicted segmentation map $m_s$. Traditional MAE-based methods typically mask image patches randomly, and then use visible patches to recover the masked patches. However, most of the forged regions are concentrated around the facial area. Random masking has high propensity to preserve some strong forged facial components that can mislead the target face recovery process. We mitigate this problem by using the predicted $m_s$ to select image patches to be masked. As demonstrated in Algorithm \ref{alg:Semantic-guide}, we first segment $m_s$ into $n$ patches, each of size $\rho$. Each patch is weighted by its average value. For patches with weights less than 0.5, their weights are replaced by random numbers between 0 and 0.1. Then, the patch weights are sorted in descending order. The top $m$ highest weight patches are selected, where the number of patches to be masked $m$ is determined by the mask ratio $\lambda$. This strategy gradually shifts the patch masking from random to source-related areas as the accuracy of the predicted segmentation map improves. Eventually, all patches related to the source face will be masked and inpainted by using other visible patches to restore the target face.

\begin{algorithm}[t]

	\caption{Semantic-Guided Masking Algorithm}
	\begin{algorithmic}[1]
		\REQUIRE Segmentation map $m_s$, patch size $\rho$, mask ratio $\lambda$
		\ENSURE{Indices of the masked and unmasked patches.}
		% \REQUIRE{$y$}
		\STATE Divide $m_s$ into $n$ patches of size $\rho$: $P \gets \text{Divide}(m_s, \rho)$;
		\STATE Calculate mean value of $m_s$: $\mu \gets \text{mean}(m_s)$;
		\STATE Calculate mean value of each patch: $\omega_i \gets \text{mean}(P_i)$;
		\STATE Generate random vector of elements $\in [0, 0.1]$: $\nu \gets \text{RandVector}(n)$;
		\IF{$\mu \geq 0.9$}
		\STATE $\omega \gets \nu$
		\ENDIF
		\FOR{$i=1$ to $n$}
		\IF{$\omega_i < 0.5$}{
			\STATE $\omega_i \gets \nu_i$
		}
		\ENDIF
		\ENDFOR

		\STATE Get sorted indices of patch weights: $\iota \gets \text{argsort}(\omega)$
		\STATE Calculate number of masked patches: $m \gets n \times \lambda$

		\STATE Get indices of masked patches: $\iota_m \gets Top_{m}(\iota)$
		\STATE Get indices of patches to keep: $\iota_k \gets Last_{n-m}(\iota)$
		% \RETURN $\iota_m, \iota_k$
	\end{algorithmic}
	\label{alg:Semantic-guide}

\end{algorithm}

\subsubsection{Identity Recovery} The aim of this step is to restore the masked patches to the corresponding patches in the target face. We utilize the background contextual information and the latent target identity information to achieve this objective.
Background context information refers to the unmasked patches within the input image, where structural and color correlations among various image components exist. These correlations between pixels or regions can assist in predicting partial information about the masked areas from the unmasked patches. The target identity information is the latent features $f_{t}$ extracted by the SIRM to guide the reconstruction of the target face. To synergize the background context information with the latent target identity information, we construct an Identity Fusion Decoder in the MAE. The target identity features and visible patch embeddings are concatenated to predict the masked patches. Its principal operations are delineated as follows.

The input image $I$ is first tokenized into a sequence of $N$ tokens $\{x_i^s\}_{i=1}^N$, where $N$ denotes the total number of image patches. For the masked version of $I$, the visible patch tokens are represented as ${x^{sv}}$. Similarly, the target face image $I_{tgt}$ is also tokenized into a sequence of $N$ image patch tokens $\{x_i^t\}_{i=1}^N$.
Following the same MAE settings as \cite{he2022masked}, we utilize a ViT as the encoder $G_{enc}^t$, and incorporate multiple cascaded transformer blocks as the Identity Fusion Decoder $G_{dec}^t$. The visible tokens $x^{sv}$ are encoded into token embeddings using linear transformation $l$, followed by the addition of positional encoding $p^{sv}$. We feed the fused embeddings into the ViT to obtain the embedding features, which are then concatenated with $f_{t}$ to generate the token embeddings with identity features. The fused features are fed into the Identity Fusion Decoder to predict the pixels of the masked patches.
\begin{align}
	x^{recm}=G_{dec}^t(G_{enc}^t(l(x^{sv})+p^{sv}) \mathbin\Vert f_{t})
\end{align}

\noindent \textbf{Identity Loss.} This loss is computed by subjecting the original and recovered faces to the FaceNet model~\cite{schroff2015facenet}  pretrained on the VGGFace2 dataset~\cite{cao2018vggface2} to extract their multi-layer identity features. The identity losses for the source and target recovery are calculated as follows:
	\begin{align}
		\mathcal{L}_{srcid}=\sum_{i = 1}^{4} ||E_{id}^i(I_{src})-E_{id}^i(I_{src}^{rec}) ||_2,\\
		\mathcal{L}_{tgtid}=\sum_{i = 1}^{4} ||E_{id}^i(I_{tgt})-E_{id}^i(I^{rec}_{tgt}) ||_2,
	\end{align}

\noindent where $E_{id}^i(\cdot)$ denotes the $i$-th layer features of FaceNet. The four layers of features are extracted from $Mixed\_6a$, $repeat\_3$, $block8$, and $last\_bn$. $I_{src}$ and $I_{tgt}$ represent the original source and target faces, respectively while $I_{src}^{rec}$ and $I_{tgt}^{rec}$ represent the reconstructed source and target faces, respectively. The overall identity loss is:
\begin{align}
	\mathcal{L}_{id}=\mathcal{L}_{srcid}+ \alpha \mathcal{L}_{tgtid},
\end{align}

\noindent \textbf{Perceptual Loss:} The perceptual loss between the original and recovered  faces is computed using the VGG19 model pretrained on the ImageNet~\cite{deng2009imagenet} dataset as follows:
	\begin{align}
		\mathcal{L}_{srcperc}=\sum_{i = 1}^{5} ||\psi _i(I_{src})-\psi_i(I_{src}^{rec}) ||_2,\\
		\mathcal{L}_{tgtperc}=\sum_{i = 1}^{5} ||\psi _i(I_{tgt})-\psi_i(I_{tgt}^{rec}) ||_2,
	\end{align}

\noindent where $\psi_i(\cdot)$ denotes the $i$-th layer features of VGG19, and these five layers are $relu1\_2$, $relu2\_2$, $relu3\_4$, $relu4\_4$, and $relu5\_4$.
The overall perceptual loss is:
\begin{align}
	\mathcal{L}_{perc}=\mathcal{L}_{srcperc}+ \beta \mathcal{L}_{tgtperc},
\end{align}

\noindent \textbf{Attribute Loss.} We utilize the attribute extraction model~\cite{he2017adaptively} pretrained on the CelebA dataset~\cite{liu2015faceattributes} to extract the facial attributes of both the recovered and pristine target faces, and use L1 smooth distance to measure their mismatch.
\begin{align}\mathcal{L}_{attr}=\left\{\begin{aligned}
		0.5(\mathcal{A}_{tgt}-\mathcal{A}_{rec})^2, \ \ \   & if\ |\mathcal{A}_{tgt}-\mathcal{A}_{rec}|<1 \\
		|\mathcal{A}_{tgt}-\mathcal{A}_{rec}|-0.5,   \ \ \  & otherwise
	\end{aligned}\right., \end{align}

\noindent where $\mathcal{A}_{tgt}$ and $\mathcal{A}_{rec}$ denote the extracted attributes of the pristine target face and the recovered target face, respectively.

\noindent \textbf{Patch Recovery Loss.}
Unlike the traditional objective of image reconstruction with the MAE, the TIRM aims to recover the target face. This is achieved by calculating the L2 distance between the recovered patches and the corresponding patches of the target face image.
	\begin{align}
		\mathcal{L}_{patch}=\sum_{i = 1}^{N_p} \left\lVert x_{i}^{tm}-x_{i}^{recm}\right\rVert_2,
	\end{align}

\noindent where $x_{i}^{tm}$ and $x_{i}^{recm}$ represent the $i$-th masked patches of the pristine target face and the recovered target face, respectively. $N_p$ denotes the number of masked patches.

\noindent \textbf{Overall Loss.} The overall loss of DFREC can be summarized as follows.
\begin{align}                                                        
	 & \mathcal{L}=\mathcal{L}_{id}+\lambda_1\mathcal{L}_{perc}+\lambda_2\mathcal{L}_{attr}+\lambda_3\mathcal{L}_{patch},
\end{align}

\noindent where $\lambda_1$, $\lambda_2$, and $\lambda_3$ are the weight parameters for balancing multiple loss functions.

\section{Experiments}

In this section, we first compare the proposed DFREC scheme with the existing SOTA approaches, and then perform an ablation study to demonstrate the effects of each module.

\subsection{Experimental Settings}

\subsubsection{Datasets} We utilize three deepfake datasets to validate the performance of our scheme:
\begin{enumerate}[\hspace*{\parindent}]
	\item[-]Faceforensics++ \cite{rossler2019faceforensics}: This dataset comprises 1000 segments of real videos, primarily centered around news broadcasts. We select two algorithms, namely DeepFake (DF) and FaceShifter (FShi), which contain a total of 2000 fake videos. Additionally, the real videos are processed using the LCR algorithm \cite{zhu2021megafs} to generate another set of 1000 fake videos. Following the splitting strategy outlined in~\cite{rossler2019faceforensics}, we obtain 100,000 training images, 20,000 validation images, and 20,000 testing images for each forgery method.

	\item[-]CelebaMegaFS~\cite{zhu2021megafs}: This dataset is constructed from the Celeba-HQ dataset \cite{karras2018progressive} for face swapping. It incorporates three deepfake methods: IDInjection (IDI), FTM, and LCR, resulting in 30,038 FTM images, 30,441 IDInjection images, and 30,010 LCR images. Each dataset is divided into training, validation, and test sets according to an 8:1:1 ratio.

	\item[-]FFHQ-E4S~\cite{liu2023fine}: This dataset is constructed based on the FFHQ dataset \cite{karras2019style}. We randomly select 10,000 images from FFHQ and use E4S \cite{liu2023fine} to generate 10 swapped images per original image. These images are distributed across training, validation, and test sets in an 8:1:1 ratio based on face identities.
\end{enumerate}

\subsubsection{Compared approaches} We compare DFREC with the following SOTA methods:
\begin{enumerate}[\hspace*{\parindent}]
	\item[-]MAT \cite{li2022mat}: It is a mask-aware transformer for large hole inpainting, which combines transformers and convolutions to reconstruct high resolution images.
	\item[-]RePaint \cite{lugmayr2022repaint}: It is a face inpainting method which uses a pretrained unconditional Denoising Diffusion Probabilistic Model (DDPM) as the generative prior.
	\item[-] RECCE \cite{cao2022end}: It is a reconstruction-based deepfake detection method, which exploits the reconstruction difference as guidance of forgery traces.
		\item[-]Delocate \cite{hu2024delocate}: It is  a deepfake face recovery approach that utilizes MAE to learn the distribution of pristine faces for the recovery of manipulated faces.
		\item[-]DFI \cite{ai2023deepfake}: It is a deepfake  inversion method that learns the reverse mapping from the forgery face to the original face.

\end{enumerate}
\begin{table*}[]\large
	\renewcommand\arraystretch{1.2}
	\caption{Performance comparison of the proposed DFREC with SOTA inpainting and deepfake recovery methods against different face-swapping forgeries on FaceForensics++, CelebaMegaFS and FFHQ-E4S datasets.}
	\label{tab:recovery}
	\resizebox{\textwidth}{!}{
		\begin{tabular}{@{}c||cccccc||cccccc||cc@{}}
			\toprule
			\multirow{3}{*}{Method}           & \multicolumn{6}{c||}{FaceForensic++} & \multicolumn{6}{c||}{CelebaMegaFS} & \multicolumn{2}{c}{FFHQ-E4S}                                                                                                                                                                                                                                             \\ \cmidrule(l){2-15}
			                                  & \multicolumn{2}{c}{DeepFake}         & \multicolumn{2}{c}{FaceShifter} & \multicolumn{2}{c||}{LCR}    & \multicolumn{2}{c}{FTM} & \multicolumn{2}{c}{IDInjection} & \multicolumn{2}{c||}{LCR} & \multicolumn{2}{c}{E4S}                                                                                                                           \\ \cmidrule(l){2-15}
			                                  & FID $\downarrow $                                 & IDSim $\uparrow $                           & FID $\downarrow $                          & IDSim $\uparrow $                   & FID $\downarrow $                             & IDSim $\uparrow $                     & FID $\downarrow $                     & IDSim $\uparrow $           & FID $\downarrow $            & IDSim           & FID $\downarrow $            & IDSim $\uparrow $           & FID $\downarrow $           & IDSim $\uparrow $           \\ \midrule
			None                              & 11.35                                & 0.4238                          & 8.43                         & 0.4252                  & 22.09                           & 0.3516                    & 11.60                   & 0.2792          & 10.83          & 0.3150          & 12.58          & 0.2318          & 8.04          & 0.2812          \\ \hline
			MAT \cite{li2022mat}               & 21.01                                & 0.4372                          & 19.23                        & 0.3874                  & \textbf{26.98}                           & 0.4026                    & 42.06                   & 0.3928          & 34.64          & 0.4203          & 35.98          & 0.3506          & 9.82          & 0.2797          \\
			Repaint \cite{lugmayr2022repaint} & 21.44                                & 0.3832                          & 19.49                        & 0.3366                  & 27.52                           & 0.3639                    & 41.18                   & 0.3874          & 33.97          & 0.3893          & 35.11          & 0.3423          & 38.97         & 0.3946          \\
			RECCE \cite{cao2022end}           & 87.28                                & 0.3755                          & 65.72                        & 0.4400                  & 89.33                           & 0.4086                    & 90.36                   & 0.2547          & 115.35         & 0.2957          & 110.69         & 0.2313          & 81.82         & 0.2668          \\
			Delocate \cite{hu2024delocate}          & 18.16                                & 0.4749                          & 37.22                        & 0.4551                  & 135.93                          & 0.3858                    & 40.75                   & 0.4201          & 39.90          & 0.4650          & 41.76          & 0.3466          & 36.88         & 0.3064          \\
			DFI \cite{ai2023deepfake}                    & 42.17             & 0.4706            & 30.39             & \textbf{0.5207}            & 37.03             & 0.4860            & 22.19             & 0.4137            & 22.20             & 0.4346          & 29.30             & 0.3442            & 29.86             & 0.3805            \\
Ours                    & \textbf{17.22}    & \textbf{0.5367}   & \textbf{18.48}    & 0.5047   & 27.92    & \textbf{0.5135}   & \textbf{15.35}    & \textbf{0.7192}   & \textbf{13.78}    & \textbf{0.7582} & \textbf{14.75}    & \textbf{0.6967}   & \textbf{8.34}     & \textbf{0.5870}  \\ \bottomrule
		\end{tabular}}
\end{table*}

\subsubsection{Evaluation Metrics}
\label{sec:evaluation_metrics}
We utilize the Frechet Inception Distance (FID) and Identity Similarity (IDSim) metrics to assess the identity recovery performance. FID is commonly employed for evaluating image similarity, while IDSim is specifically designed to assess the identity similarity between the reconstructed image and the original image \cite{shiohara2023blendface}. To compute IDSim, we leverage the pretrained FaceNet~\cite{schroff2015facenet} model to extract identity features from both the original and  recovered target faces, followed by computing the cosine similarity between these extracted features. Attribute preservation is measured by computing cosine similarity between adaptive attribute embeddings \cite{liu2015faceattributes}. Pose consistency is evaluated using pose parameters from a 3D face reconstruction model \cite{deng2019accurate} with angular distance measurement.
Additionally, we use  $Acc_{id}$ to represent the identity-based recovery accuracy. It is computed based on the identity similarity between the input face and the recovered target face. A high identity similarity implies a small discrepancy, and vice versa. If the identity similarity surpasses the threshold of 0.9, the input face is categorized as real. Otherwise, the input face is classified as fake.

\subsubsection{Implementation Details} We employ an EfficientNet-B0 as the backbone for the ISM and a ViT16 as the backbone for the MAE network. All the face images are extracted by RetinaFace and then resized to 224 $\times$ 224 pixels. During training, the models are simultaneously trained on both real and forgery images. Additionally, we utilize the algorithm from BI \cite{li2020face} to continuously generate forgery faces and masks, which aids in enhancing the segmentation performance of the ISM. The source and target faces recovered from the real images are the real images themselves. For the forgery images, the recovered faces correspond to the source and target faces present in the dataset. We train our model using four NVIDIA 4090 GPUs. The mask ratio used in the TIRM is set as 0.5 and the patch size is set as 16. We optimize the models using the AdamW optimizer with a learning rate of 1e$-$4 and a weight decay of 1e$-$5. To improve the target face recovery quality, the parameters $\alpha$ and $\beta$ are set as 2. The loss weights $\lambda_1$, $\lambda_2$, and $\lambda_3$ are set as 0.5.

\subsection{Intra-dataset Deepfake Recovery}

We evaluate the identity recovery performance on the FaceForensics++, CelebaMegaFS and FFHQ-E4S datasets. Forgery images in the test sets are fed into DFREC to generate the recovered source and target faces. Quantitative and qualitative comparisons are made between DFREC and other face inpainting and deepfake recovery methods.

\subsubsection{Quantitative Evaluation}
We use FID to assess the image recovery quality, and IDSim to evaluate identity recovery quality between the recovered and the original target face for comparison with other methods. We leverage publicly available pre-trained models or existing codes for MAT, Repaint, and RECCE. Delocate and DFI are implemented by following the detailed methodologies presented in the corresponding papers. For face inpainting methods like MAT and Repaint, we employ dlib for the facial detection, and then mask areas that contain facial components before inpainting. Table \ref{tab:recovery} presents the recovery performance across the three datasets. The `None' method in the first line shows the FID and IDSim scores between the forgery face and the original target face, and the next six lines show the values of the corresponding metrics between the recovered target face and the original target face. We analyze the results of image recovery on the FaceForensics++, CelebaMegaFS, and FFHQ-E4S datasets.  Compared with existing face inpainting methods \cite{li2022mat,lugmayr2022repaint} and deepfake recovery methods \cite{cao2022end, hu2024delocate,ai2023deepfake}, DFREC achieves the best FID scores and identity similarity in different types of face swapping forgeries, demonstrating a higher capability of recovering images that are closer to the original target face identity. 
\begin{figure*}[t]
	\centering
	\includegraphics[width=\linewidth]{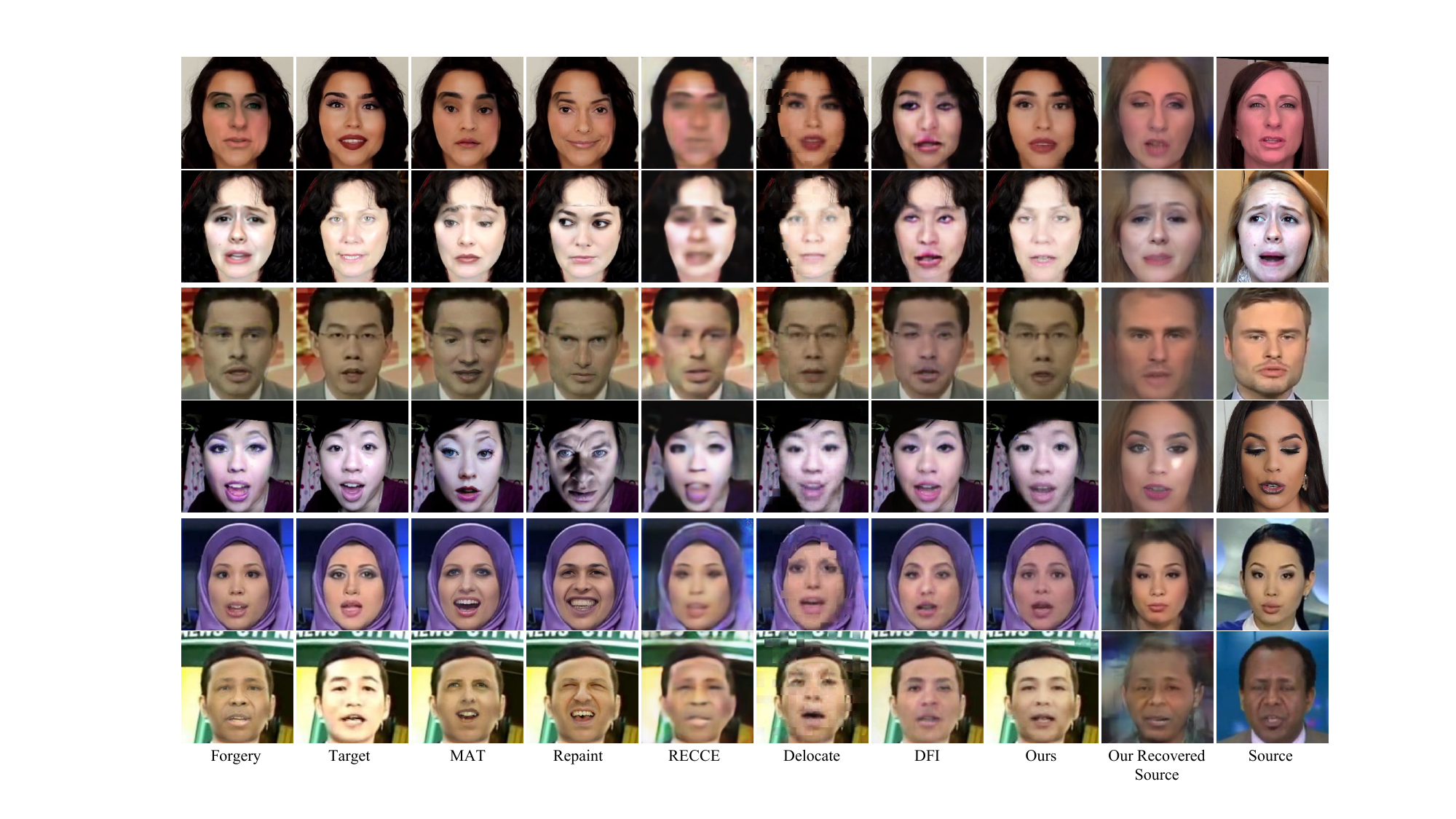}
	\caption{Comparison of identity recovery quality of different face inpainting and deepfake recovery methods on face-swapping images of the FaceForensics++ dataset.}
	\label{fig:ffcompare}
	
\end{figure*}

\begin{figure*}[t]
	\centering
	\includegraphics[width=\linewidth]{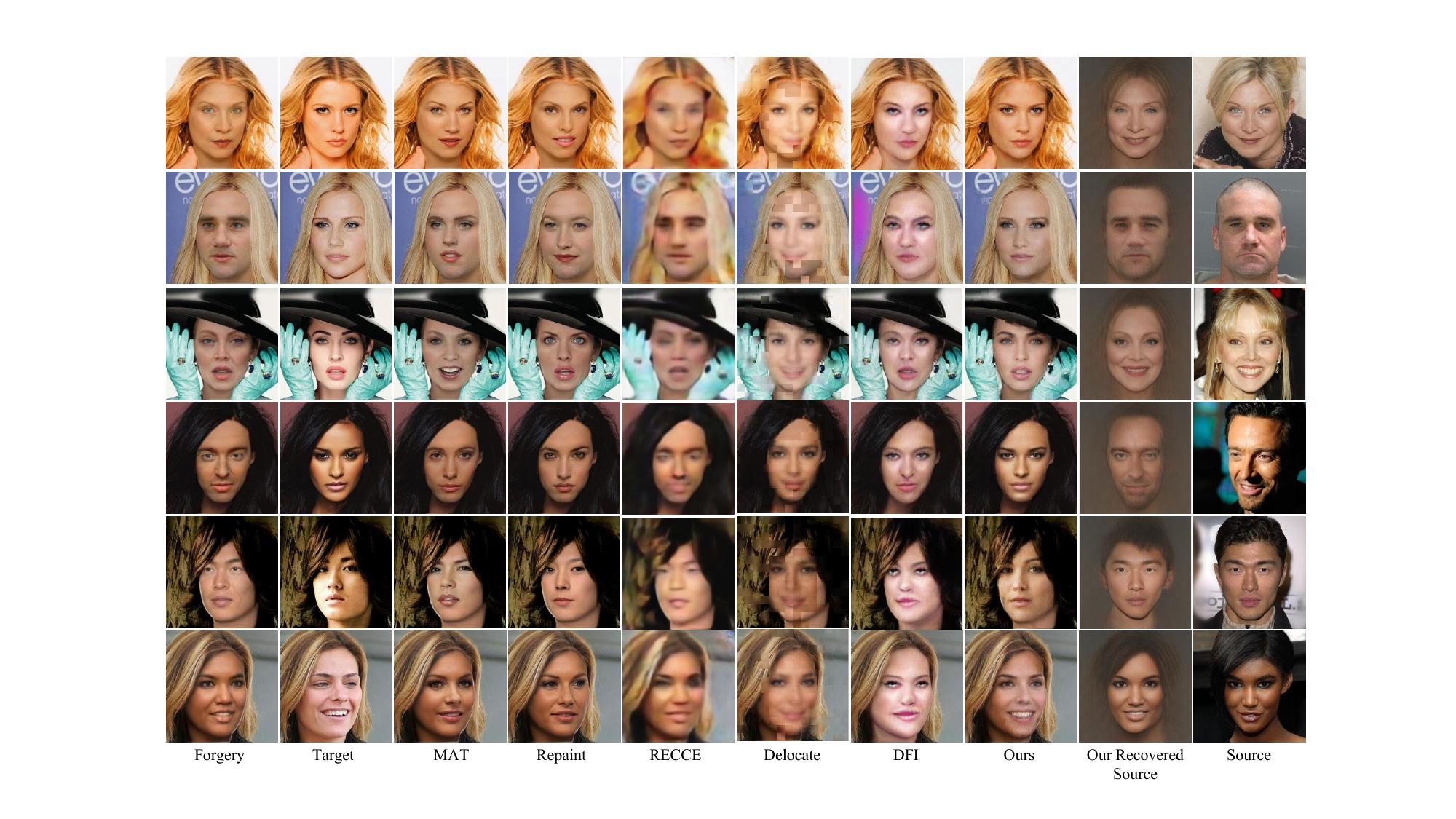}
	\caption{Comparison of identity recovery quality of different face inpainting and deepfake recovery methods on face-swapping images of the CelebaMegaFS dataset.}
	\label{fig:celebcompare}
\end{figure*}

\begin{figure*}[!h]
	\centering
	\includegraphics[width=\linewidth]{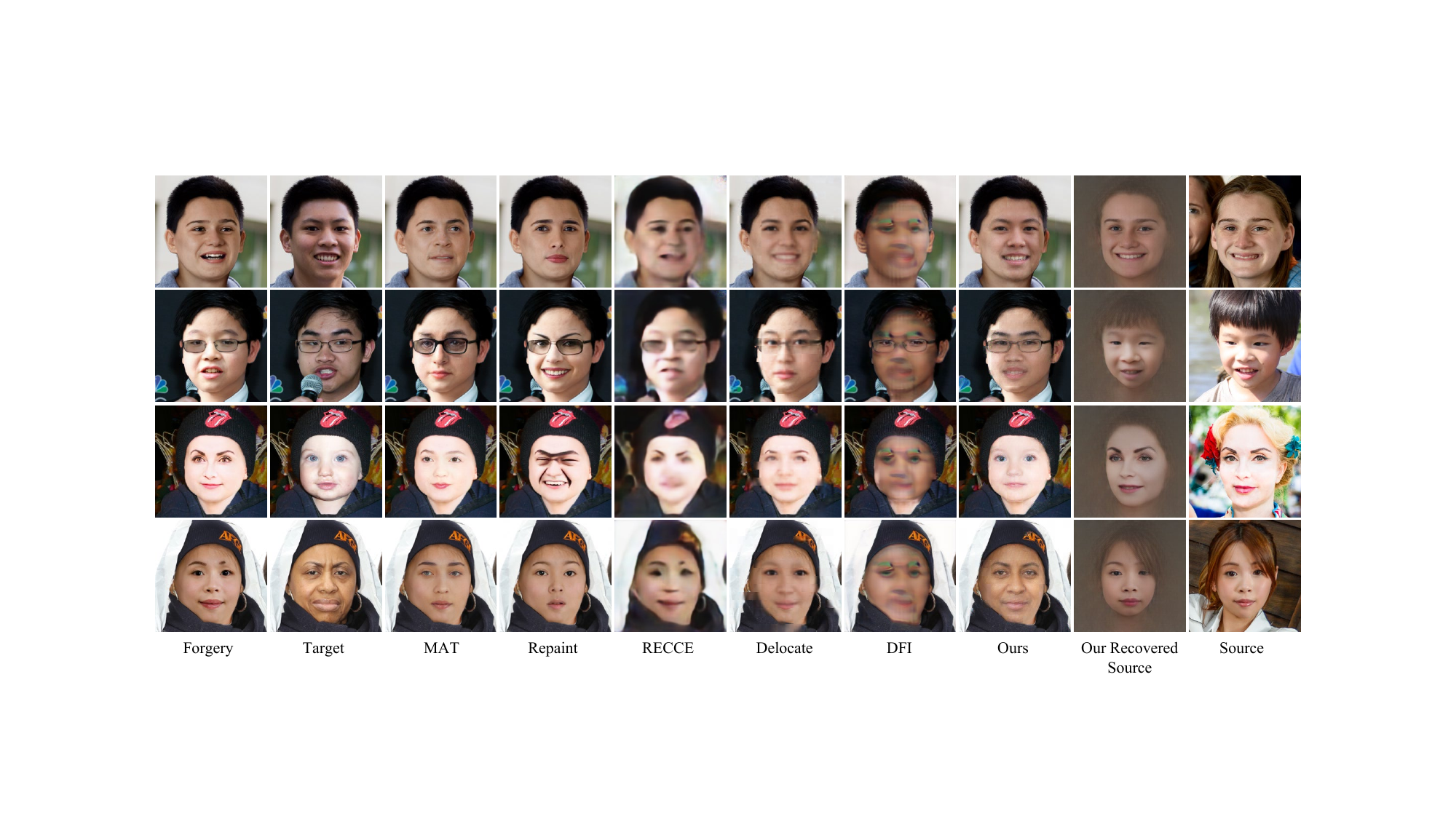}
	\caption{Comparison of identity recovery quality of different face inpainting and deepfake recovery methods on face-swapping images of the FFHQ-E4S dataset.}
	\label{fig:ffhqcompare}
\end{figure*}

\subsubsection{Qualitative Evaluation}
Figs \ref{fig:ffcompare} - \ref{fig:ffhqcompare} compare the target identity recovery quality between DFREC  and several SOTA face inpainting and deepfake recovery methods.  The evaluations were conducted on the test sets of the FaceForensics++, CelebaMegaFS, and FFHQ-E4S, respectively. As shown in Figs \ref{fig:ffcompare} - \ref{fig:ffhqcompare}, although MAT and Repaint can synthesize realistic facial images based on the unmasked region, there are still identity gaps between the synthesized images and the original target faces. We also compared DFREC with three deepfake recovery approaches \cite{cao2022end, hu2024delocate,ai2023deepfake}. They tend to generate blurred and unidentified faces, which are apparently  different from the original target faces. In contrast, DFREC can synthesize more realistic target faces with similar identities as the original target faces.

\subsection{Generalization of Deepfake Recovery}
\begin{figure*}[t]
	\centering
	\includegraphics[width=\linewidth]{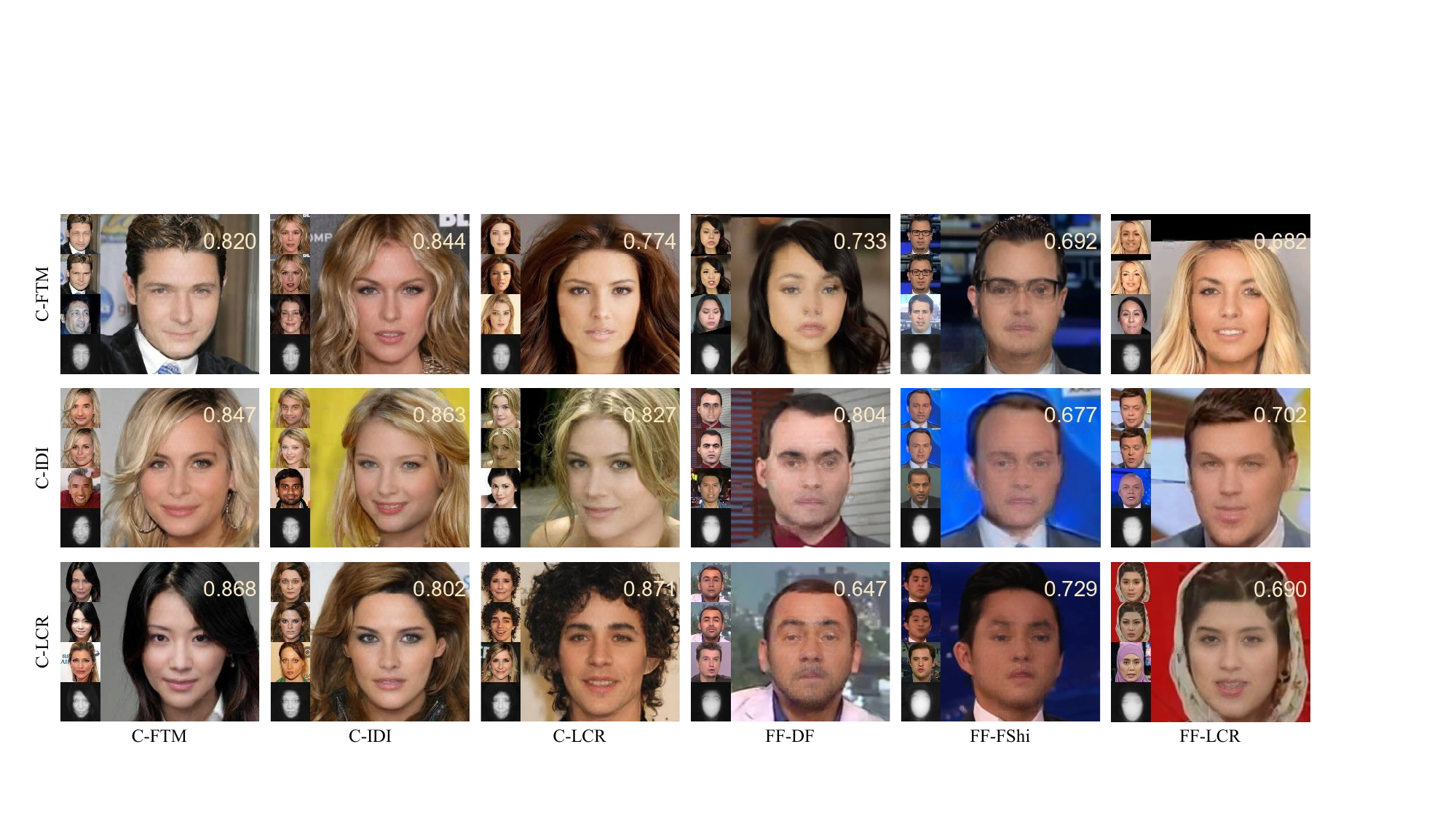}
	\caption{Target face recovery on unseen forgeries. The vertical axis represents the training dataset, and the horizontal axis denotes the evaluation dataset. From top to bottom of the inset on the left side of the recovered images are the forged face, target face, source face, and segmentation map. The value at the top-right corner indicates the identity similarity between the recovered and target faces. }
	\label{fig:generalization}
\end{figure*}

In this section, we evaluate the  deepfake recovery performance of DFREC on unseen forgeries. To visually demonstrate the generalization effect of DFREC, we present the recovered target images across different datasets in Fig. \ref{fig:generalization}. DFREC are trained on the various forgery types of CelebaMegaFS dataset and then evaluated across FaceForensics++ and CelebaMegaFS datasets.  It can be seen that DFREC could well recover the target faces from the  unseen forgery images. 
Also, we calculated the identity between the recovered target faces and the original target faces. DFREC is trained on a specific forgery set and then evaluated across other forgeries.  Table \ref{tab:generalization_idsim} presents the identity similarity $\emph{IDSim}$ on various types of forgery data. It can be observed that DFREC generally outperforms DFI \cite{ai2023deepfake} in terms of the generalized deepfake recovery. 

\begin{table}[tbp]
    \caption{Generalization of deepfake recovery compared with DFI \cite{ai2023deepfake} across diverse forgery images in the CelebaMegaFS dataset. $IDSim$ is used as the evaluation metric. Intra-dataset evaluations are highlighted in red. }
    \renewcommand\arraystretch{1.2}
    \label{tab:generalization_idsim}
    \resizebox{\linewidth}{!}{
    \setlength{\tabcolsep}{4mm}{
    \begin{tabular}{|c|c|c|c|c|}
    \hline
    Methods & Train                 & FTM                                    & IDI                                    & LCR                                    \\ \hline
    DFI \cite{ai2023deepfake}    &                       & \textcolor{red}{ 0.4137}          & 0.4318                                 & 0.3098                                 \\ \cline{1-1} \cline{3-5} 
    DFREC  & \multirow{-2}{*}{FTM} & \textcolor{red}{ \textbf{0.7192}} & \textbf{0.6586}                        & \textbf{0.4711}                        \\ \hline
    DFI \cite{ai2023deepfake}    &                       & 0.3938                                 & \textcolor{red}{ 0.4346}          & 0.2966                                 \\ \cline{1-1} \cline{3-5} 
    DFREC  & \multirow{-2}{*}{IDI} & \textbf{0.6203}                        & \textcolor{red}{ \textbf{0.7582}} & \textbf{0.4546}                        \\ \hline
    DFI \cite{ai2023deepfake}    &                       & 0.3603                                 & 0.3802                                 & \textcolor{red}{ 0.3442}          \\ \cline{1-1} \cline{3-5} 
    DFREC  & \multirow{-2}{*}{LCR} & \textbf{0.6332}                        & \textbf{0.6843}                        & \textcolor{red}{ \textbf{0.6967}} \\ \hline
    \end{tabular}}}
    \end{table}

\subsection{Deepfake Recovery Accuracy}
\begin{figure}[t]
	\centering
	\includegraphics[width=\linewidth]{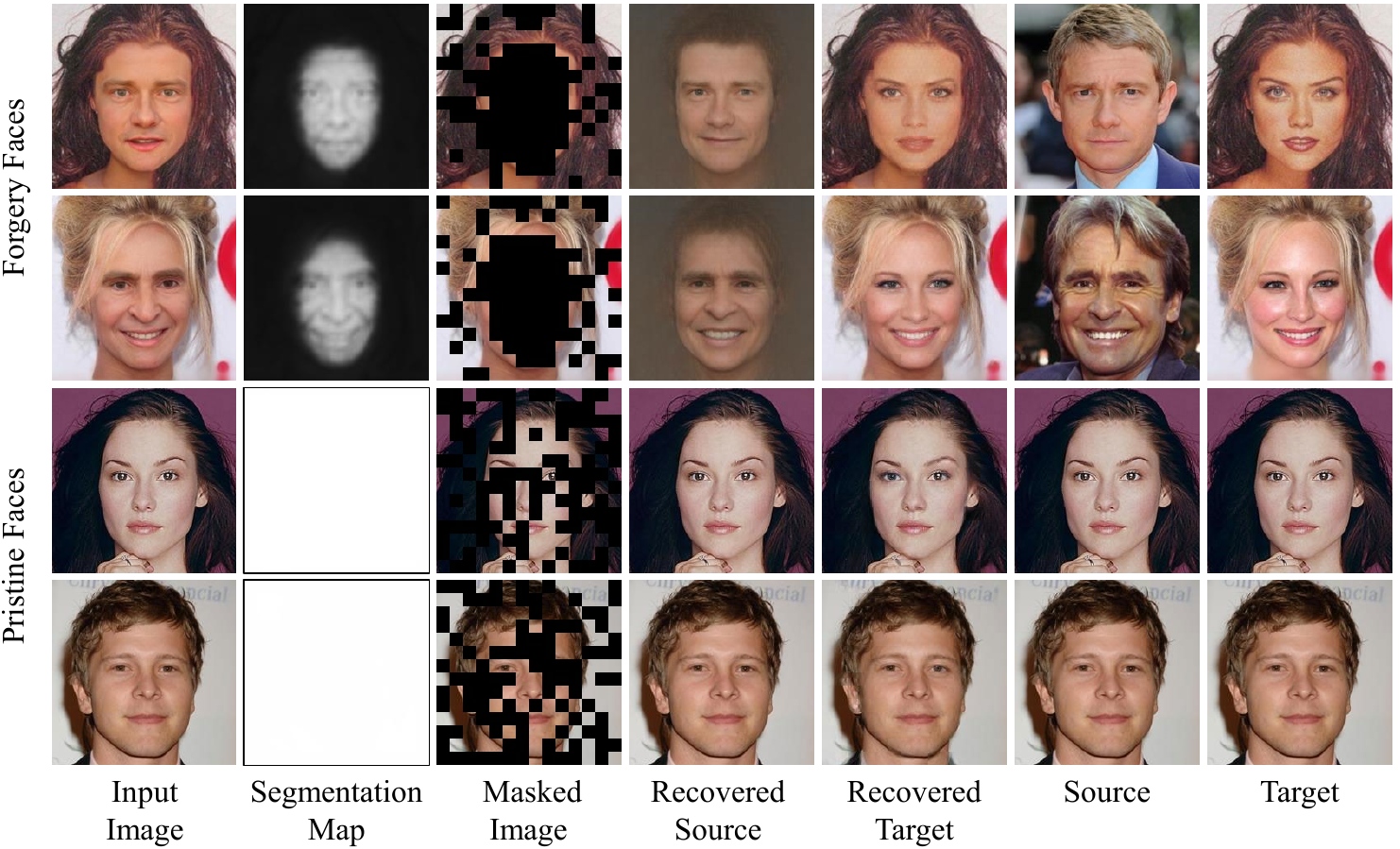}
	\caption{Intermediate outputs of DFREC alongside the reconstructed source and target faces. The segmentation map $m_s$ is generated by the ISM, and the masked image is produced by applying $m_s$ to the input image. Our DFREC could effectively distinguishes between forgery and pristine regions, enabling precise recovery of both source and target faces.}

	\label{fig:dfrecovery}
\end{figure}

The proposed DFREC should exhibit distinct behaviors when processing manipulated versus authentic images. As illustrated in Fig. \ref{fig:dfrecovery}, DFREC generates different output results when presented with forgery and pristine facial images. Specifically, for forgery faces, DFREC precisely reconstructs both the source and target faces from the manipulated input. In contrast, for pristine faces, DFREC ensures that the recovered source and target faces remain identical to the original input. To further evaluate the effectiveness of our approach, we calculated the recovery accuracy ($Acc_{id}$) under intra-dataset and cross-dataset settings.

\subsubsection{Intra-dataset Recovery Accuracy}
We first assess the intra-dataset recovery accuracy of DFREC on the FaceForensics++, CelebaMegaFS, and FFHQ-E4S datasets. The identity similarity between the recovered target image and the forged image should exhibit significant differences, whereas for authentic images, the similarity should remain consistent. Table \ref{tab:detection} presents the identity-based detection accuracy $Acc_{id}$ on various types of forgery and pristine images. We compare DFREC with existing deepfake recovery methods \cite{cao2022end,hu2024delocate,ai2023deepfake} across three datasets. It can be observed that DFREC generally outperforms existing deepfake recovery methods in terms of detection performance. Its face recovery accuracy is obviously higher than RECCE\cite{cao2022end}, Delocate\cite{hu2024delocate}, and DFI\cite{ai2023deepfake}.

\begin{table}[]\large
	\caption{Comparison of $Acc_{id}$ with deepfake recovery methods on the  FaceForensics++, CelebaMegaFS, and FFHQ-E4S datasets.}
	\label{tab:detection}
    \renewcommand\arraystretch{1.3}
    \resizebox{\linewidth}{!}{
	\setlength{\tabcolsep}{1.2mm}{
		\begin{tabular}{@{}clccclccclc@{}}
			\toprule
			\multirow{2}{*}{Methods}                        &  & \multicolumn{3}{c}{FaceForensic++}             &  & \multicolumn{3}{c}{CelebaMegaFS}                 &  & FFHQ-E4S       \\ \cmidrule(l){3-11} 
															&  & DF           & FShi           & LCR            &  & FTM            & IDI            & LCR            &  & E4S            \\ \cmidrule(r){1-1} \cmidrule(lr){3-5} \cmidrule(lr){7-9} \cmidrule(l){11-11} 
			RECCE \cite{cao2022end}        &  & 43.39        & 49.70          & 48.77          &  & 52.25          & 56.41          & 56.00          &  & 50.35          \\
			Delocate \cite{hu2024delocate} &  & 89.32        & 89.83          & 67.16          &  & 67.37          & 68.82          & 65.52          &  & 69.01          \\
			DFI \cite{ai2023deepfake}      &  & 90.81        & 86.50          & 94.43          &  & 96.45          & 92.70          & 98.83          &  & 97.18          \\
			DFREC                                           &  & \textbf{100} & \textbf{99.38} & \textbf{99.55} &  & \textbf{99.13} & \textbf{99.32} & \textbf{99.07} &  & \textbf{98.04} \\ \bottomrule
			\end{tabular}}}
\end{table}

\subsubsection{Cross-dataset Recovery Accuracy}
Here, we assess the generalized deepfake recovery performance of DFREC. We trained DFREC on the CelebaMegaFS dataset and then evaluate its deepfake recovery accuracy across FaceForensics++ and CelebaMegaFS datasets. Given that DFI is specifically designed for deepfake recovery and represents the best-performing existing method, we compare the deepfake recovery performance of DFREC against DFI. Table \ref{tab:generalization} presents the identity-based recovery accuracy $Acc_{id}$ on various types of forgery data. It can be observed that DFREC generally outperforms DFI \cite{ai2023deepfake} in terms of the generalized deepfake recovery.

\begin{table}[tbp]\large
	\caption{Generalization of deepfake recovery compared with DFI \cite{ai2023deepfake} across diverse forgery images. $Acc_{id}$ is used as the evaluation metric. Intra-dataset evaluations are highlighted in red.}
    \label{tab:generalization}
	\renewcommand\arraystretch{1.2}
	\resizebox{\linewidth}{!}{
		\begin{tabular}{|c|c|ccc|ccc|}
			\hline
									 &                         & \multicolumn{3}{c|}{CelebaMegaFS}                                                                                                                                  & \multicolumn{3}{c|}{FaceForensics++}                                                       \\ \cline{3-8} 
			\multirow{-2}{*}{Method} & \multirow{-2}{*}{Train} & {FTM}                                    & {IDI}                                    & LCR                                    & {DF}             & {FShi}           & LCR            \\ \hline
			DFI \cite{ai2023deepfake}                      &                         & {\textcolor{red}{75.20}}           & {70.43}                                  & 49.03                                  & {42.32}          & {33.01}          & 67.96          \\ \cline{1-1} \cline{3-8} 
			DFREC                    & \multirow{-2}{*}{C-FTM} & {\textcolor{red}{\textbf{100.00}}} & {\textbf{99.95}}                         & \textbf{98.37}                         & {\textbf{99.41}} & {\textbf{85.44}} & \textbf{83.14} \\ \hline
			DFI \cite{ai2023deepfake}                      &                         & {58.63}                                  & {\textcolor{red}{55.89}}           & 39.20                                  & {26.76}          & {19.86}          & 44.88          \\ \cline{1-1} \cline{3-8} 
			DFREC                    & \multirow{-2}{*}{C-IDI} & {\textbf{100.00}}                        & {\textcolor{red}{\textbf{100.00}}} & \textbf{98.73}                         & {\textbf{99.78}} & {\textbf{90.30}} & \textbf{94.38} \\ \hline
			DFI \cite{ai2023deepfake}                      &                         & {91.83}                                  & {89.24}                                  & \textcolor{red}{94.27}           & {75.63}          & {64.96}          & 77.03          \\ \cline{1-1} \cline{3-8} 
			DFREC                    & \multirow{-2}{*}{C-LCR} & {\textbf{100.00}}                        & {\textbf{100.00}}                        & \textcolor{red}{\textbf{100.00}} & {\textbf{99.27}} & {\textbf{91.82}} & \textbf{97.30} \\ \hline
			\end{tabular}
	}
	\end{table}

\begin{figure}[t]
	\centering
	\includegraphics[width=\linewidth]{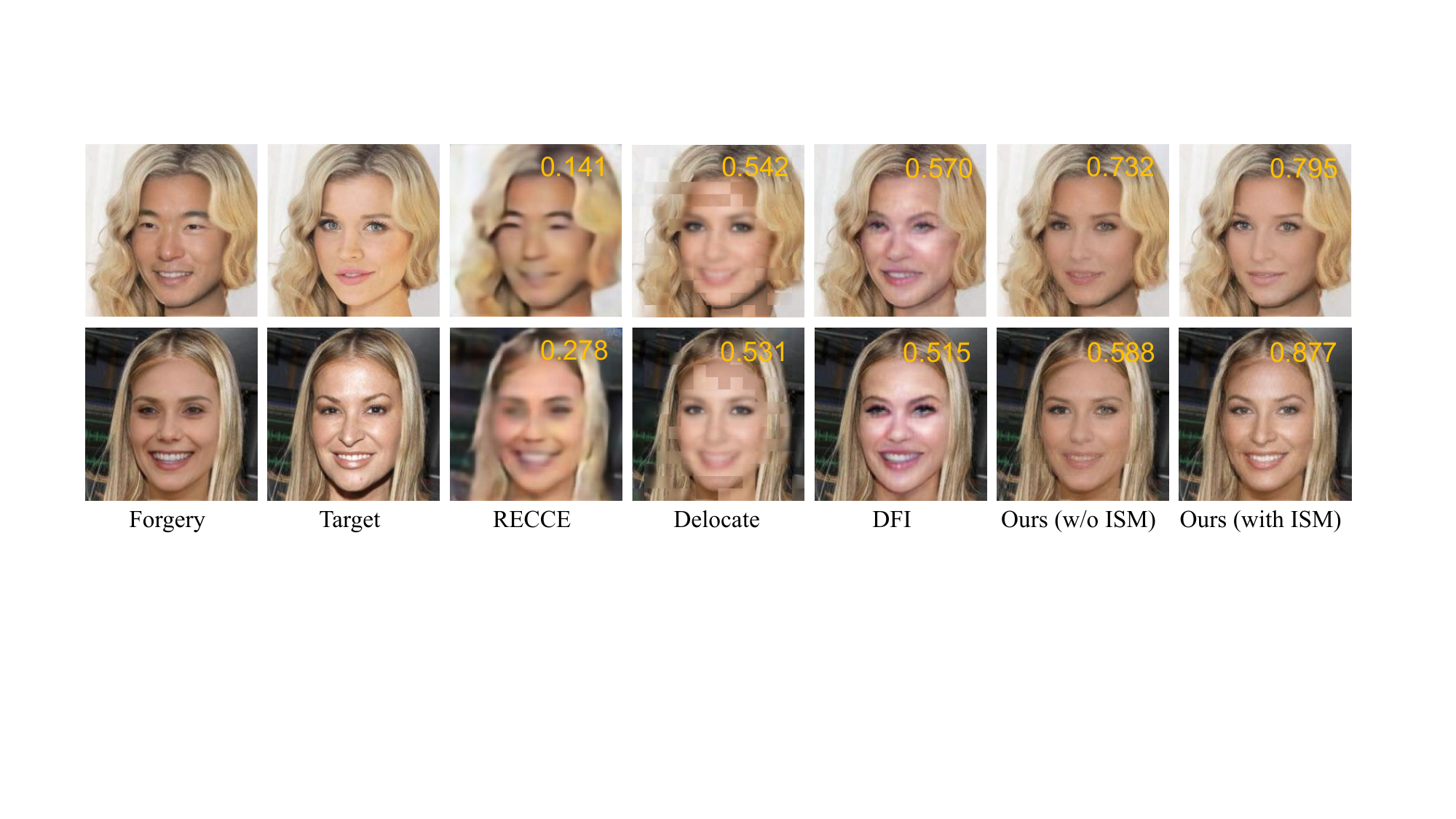}
	\caption{Ablation experiments for the ISM. The value at the top-right corner of each recovered image indicates the identity similarity between the recovered and target faces. The images are selected from the CelebaMegaFS-LCR dataset.}
	\label{fig:ISMab}
\end{figure}

\subsection{Robustness Evaluation of Deepfake Recovery}
\begin{table}[]\huge
	\renewcommand\arraystretch{1.2}
	\caption{Robustness evaluation of three forgery types (FTM, IDInjection, and LCR) in the CelebaMegaFS dataset. }
	\resizebox{\linewidth}{!}{
		\begin{tabular}{@{}cccccccccc@{}}
			\toprule
			\multirow{2}{*}{Methods} & \multicolumn{3}{c}{GaussianBlur}                    & \multicolumn{3}{c}{JPEGCompression}                 & \multicolumn{3}{c}{GaussianNoise}                   \\ \cmidrule(l){2-10} 
									 & FTM             & IDI             & LCR             & FTM             & IDI             & LCR             & FTM             & IDI             & LCR             \\ \midrule
			DFI \cite{ai2023deepfake}                      & 0.3788          & 0.4230          & 0.3294          & 0.3999          & 0.4394          & 0.3421          & 0.3864          & 0.4244          & 0.3366          \\
			DFREC                    & \textbf{0.6902} & \textbf{0.7240} & \textbf{0.6761} & \textbf{0.7055} & \textbf{0.7305} & \textbf{0.6816} & \textbf{0.7039} & \textbf{0.7324} & \textbf{0.6793} \\ \bottomrule
			\end{tabular}}

	\label{tab:robustness}
	\end{table}
We also evaluated the robustness of our approach on the CelebaMegaFS dataset. Specifically, we applied various post-processing operations to the input forgery images and subsequently measured the identity similarity between the recovered target images and the original target images, as shown in Table \ref{tab:robustness}. In this context, GB refers to Gaussian Blur (with a random kernel size ranging from 3 to 7), JPEG indicates JPEG Compression (with a random quality setting between 50 and 99), and GN represents Gaussian Noise (with variance randomly set between 10 and 50). Compared with DFI \cite{ai2023deepfake}, our method exhibits better robustness to these perturbations,  demonstrating its suitability for real-world applications.

\subsection{Ablation Study}
In this section, we present an ablation study conducted on the CelebaMegaFS dataset to validate the efficacy of the two key components of DFREC. We systematically dismantle the ISM and Target Identity Fusion to evaluate their individual contributions.

\subsubsection{Identity Segmentation Module}
To demonstrate its significance, we remove the ISM and input the image directly to the TIRM. For the MAE of TIRM, we employ a uniform mask encompassing the entire target face region. The visualization results on the CelebaMegaFS dataset are shown in Fig. \ref{fig:ISMab}. It shows that DFREC with ISM tends to generate more accurate target face. We also conducted quantitative evaluations on the three types of forgeries with CelebaMegaFS. The identity similarity between the recovered and the original target faces is calculated and shown in Table \ref{tab:ablation}. The effectiveness of the ISM is corroborated by the improved identity similarity when the ISM is included.

\subsubsection{Target Identity Fusion}
In DFREC, we hypothesize that the segmented source information contains target identity features. Thus, we use the SIRM to extract the target identity features and fuse them with the visible token embeddings obtained from the MAE to facilitate the target face recovery. To demonstrate its efficacy, we conduct an ablation study by removing the fusion strategy to solely rely on unmasked patches for recovery. As shown in Table \ref{tab:ablation}, the ability of DFREC to recover the target face is relatively weak without the fusion strategy. The improvement is significant with the fusion strategy, which attests our hypothesis that the identity fusion plays a pivotal role in  enhancing the quality of the target face recovery.

\begin{table}[t]\huge
	\caption{Ablation experiments of the ISM and Fusion strategy. The evaluations are conducted on the CelebaMegaFS images deepfaked by FTM, IDInjection and LCR.}
	\label{tab:ablation}
	\renewcommand\arraystretch{1.2}
	\resizebox{\linewidth}{!}{
		\begin{tabular}{@{}cclcclcclcc@{}}
			\toprule
			\multicolumn{2}{c}{\textbf{Settings}} &  & \multicolumn{2}{c}{FTM}             &  & \multicolumn{2}{c}{IDInjection}     &  & \multicolumn{2}{c}{LCR}             \\ \midrule
			ISM               & Fusion            &  & FID$\downarrow $ & IDSim$\uparrow $ &  & FID$\downarrow $ & IDSim$\uparrow $ &  & FID$\downarrow $ & IDSim$\uparrow $ \\ \cmidrule(r){1-2} \cmidrule(lr){4-5} \cmidrule(lr){7-8} \cmidrule(l){10-11} 
							  &                   &  & 84.97            & 0.3241           &  & 86.94            & 0.3573           &  & 80.86            & 0.2685           \\
			$\surd$           &                   &  & 22.87            & 0.5374           &  & 23.60            & 0.5520           &  & 26.52            & 0.4591           \\
							  & $\surd$           &  & 30.53            & 0.5388           &  & 29.09            & 0.5436           &  & 32.19            & 0.5687           \\
			$\surd$           & $\surd$           &  & \textbf{15.35}   & \textbf{0.7192}  &  & \textbf{13.78}   & \textbf{0.7582}  &  & \textbf{14.75}   & \textbf{0.6967}  \\ \bottomrule
			\end{tabular}
	}
	
\end{table}

\subsubsection{Effect of Each Loss}
In our approach, we incorporate identity loss ($\mathcal{L}_{id}$), perceptual loss ($\mathcal{L}_{perc}$), patch loss ($\mathcal{L}_{patch}$), and attribute loss ($\mathcal{L}_{attr}$) to ensure the quality of both the reconstructed source and target faces. To evaluate the effectiveness of each loss component, we trained and tested the model on the FTM subset of the CelebaMegaFS dataset using various loss configurations. The experimental results (Identity Similarity, Attribute Consistency, Pose Preservation, PSNR, and SSIM) are summarized in Table \ref{tab:ablationloss}. All four loss terms contribute to performance gains. We take the model using only identity loss as our baseline, which achieves satisfactory identity preservation but suffers from lower visual quality. The patch loss then refines pixel-level details, and adding perceptual loss further enhances visual metrics (PSNR, SSIM). Finally, incorporating the attribute loss improves the accuracy of facial attributes and poses (Attr, Pose). By jointly leveraging these four loss functions, our method achieves superior reconstruction of the target face, striking a favorable balance between identity fidelity and visual realism.

\begin{table}[]\Large
	\caption{Ablation experiments of each loss. The evaluations are conducted on the CelebaMegaFS images deepfaked by FTM, IDInjection and LCR.}
	\label{tab:ablationloss}
	\renewcommand\arraystretch{1.2}
	\resizebox{\linewidth}{!}{
	\begin{tabular}{@{}cccc|ccccc@{}}
	\toprule
	$\mathcal{L}_{id}$ & $\mathcal{L}_{perc}$ & $\mathcal{L}_{patch}$ & $\mathcal{L}_{attr}$ & IDSIM  & Attr  & Pose   & PSNR  & SSIM   \\ \midrule
	$\checkmark$  &       &       &           & 0.6926 & 90.14 & 130.65 & 12.23 & 0.4239 \\
	$\checkmark$  & $\checkmark$     &       &           & 0.7004 & 90.70 & 108.98 & 12.27 & 0.4262 \\
	$\checkmark$  &       & $\checkmark$     &           & 0.7072 & 94.96 & 10.34  & 21.93 & 0.6701 \\
	$\checkmark$  &       &       & $\checkmark$         & 0.6390 & 95.54 & 4.89   & 21.59 & 0.6946 \\
	$\checkmark$  & $\checkmark$     & $\checkmark$     &           & 0.6336 & 95.25 & 4.27   & 21.88 & 0.6936 \\
	$\checkmark$  & $\checkmark$     &       & $\checkmark$         & 0.6427 & 95.28 & 5.10   & 21.40 & 0.6715 \\
	$\checkmark$  &       & $\checkmark$     & $\checkmark$         & 0.7109 & 95.02 & 10.09  & 21.92 & 0.6758 \\
	$\checkmark$  & $\checkmark$     & $\checkmark$     & $\checkmark$         & 0.7192 & 96.07 & 2.62   & 22.50 & 0.7507 \\ \bottomrule
	\end{tabular}}
	\end{table}

\section{Limitations}
In this work, we introduce a deepfake face recovery method based on masked autoencoders, capable of reconstructing both the original target face and the source face from a face-swapped image. Despite its effectiveness, our approach has some limitations. First, deepfake face recovery is inherently challenging, and the success of our method relies on the presence of identity features related to the target image within the forged image. In scenarios where the forged face is not derived from the target face (e.g., through direct cropping and splicing), the performance of our approach is significantly constrained. Second, substantial variations in face images due to differing capture conditions can affect the performance of our method. This limitation can be mitigated by assembling a sufficiently diverse set of face images and employing existing deepfake face-swapping algorithms to generate synthetic data for training. Nevertheless, our proposed method can effectively recover both the original source and target faces, offering a novel tool for assisting in the collection of judicial evidence.

\section{Conclusion}

This paper proposes a novel deepfake identity recovery framework, DFREC, which is capable of restoring both the source and target faces from the deepfake faces. DFREC incorporates an ISM to effectively segment the source and target face information, a SIRM to restore the source face and extract latent target identity features, and a Mask Autoencoder-based TIRM to integrate the background information with the extracted target identity features for the target face reconstruction. We conduct extensive evaluations on deepfake images generated from the FaceForensics++, CelebaMegaFS, and FFHQ-E4S datasets by six different face swapping algorithms. The results demonstrate that DFREC outperforms existing face inpainting methods and deepfake recovery approaches, confirming its potential as a forensic tool for deepfake source and target identities tracing.

\bibliography{main}
\bibliographystyle{IEEEtran}
% \newpage

\section*{Biography Section}

\vspace{11pt}

\bf\vspace{-35pt}
\begin{IEEEbiography}[{\includegraphics[width=1in,height=1.25in,clip,keepaspectratio]{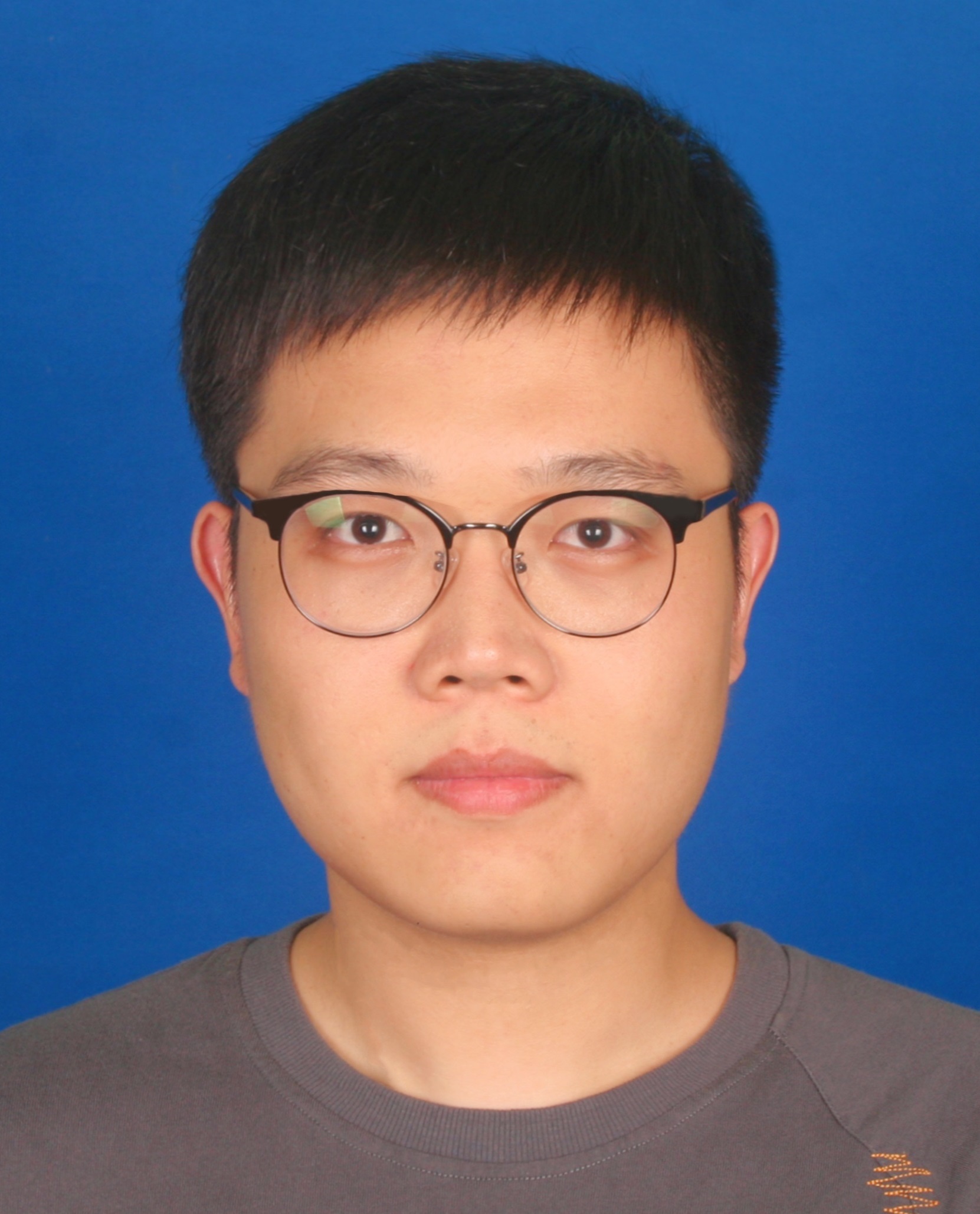}}]{Peipeng Yu} is currently pursuing Ph.D. degree in College of Information Science and Technology/Cyber Security at Jinan University, Guang Zhou, China. His research interests include artificial intelligence security and video forensics.
\end{IEEEbiography}


\bf\vspace{-33pt}
\begin{IEEEbiography}[{\includegraphics[width=1in,height=1.25in,clip,keepaspectratio]{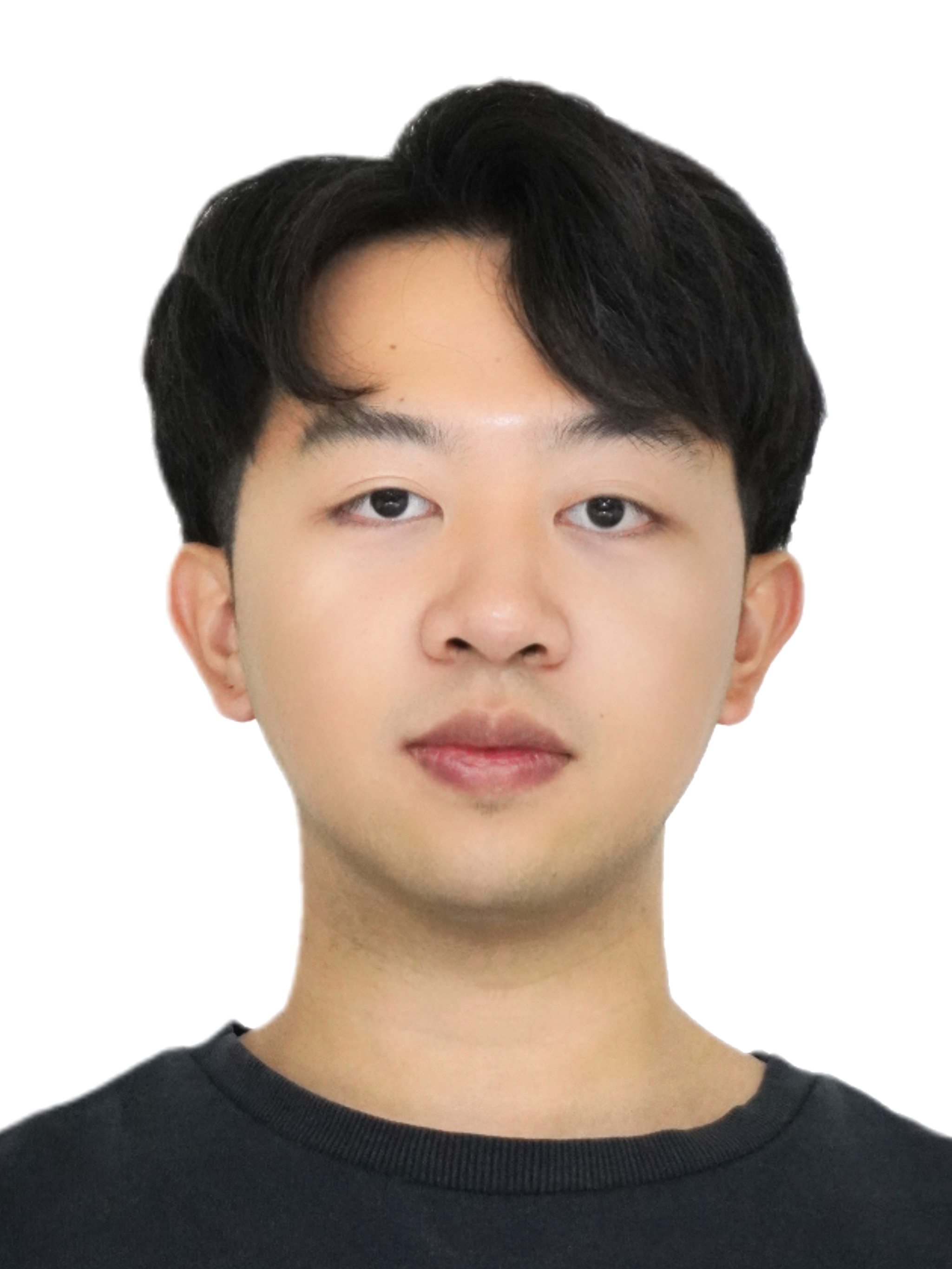}}]{Hui Gao} is currently pursuing master's degree in College of Cyber Security at Jinan University, Guang Zhou, China. His research interests include deepfake detection and image forensics.
\end{IEEEbiography}

\vspace{11pt}

\bf\vspace{-33pt}
\begin{IEEEbiography}[{\includegraphics[width=1in,height=1.25in,clip,keepaspectratio]{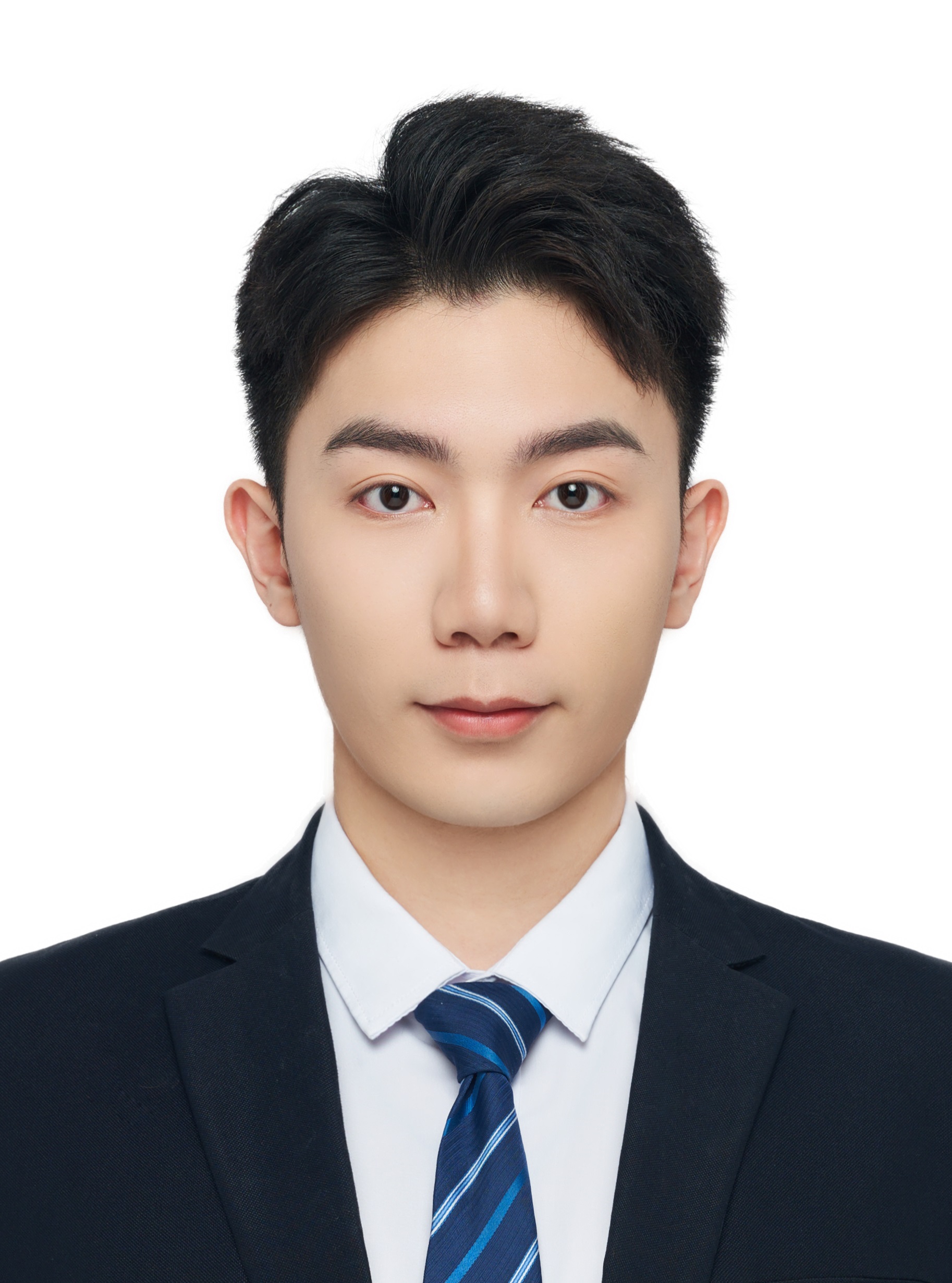}}]{Zhitao Huang} is currently pursuing master's degree in College of Information Science and Technology/Cyber Security at Jinan University, Guang Zhou, China. His research interests include artificial intelligence security and backdoor attacks.
\end{IEEEbiography}

\vspace{11pt}

\bf\vspace{-33pt}
\begin{IEEEbiography}[{\includegraphics[width=1in,height=1.25in,clip,keepaspectratio]{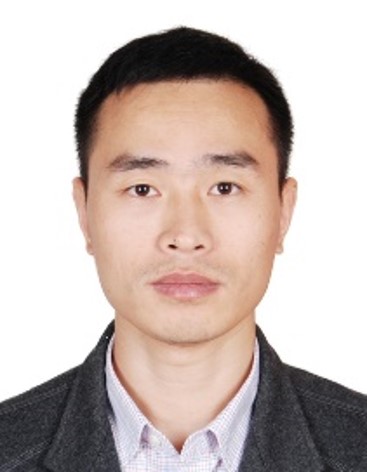}}]{Zhihua Xia} received his Ph.D. degree in computer science and technology from Hunan University, China, in 2011, and worked successively as a lecturer, an associate professor, and a professor with College of Computer and Software, Nanjing University of Information Science and Technology. He is currently a professor with the College of Cyber Security, Jinan University, China. He was a visiting scholar at New Jersey Institute of Technology, USA, in 2015, and was a visiting professor at Sungkyunkwan University, Korea, in 2016. He serves as a managing editor for IJAACS. His research interests include AI security, cloud computing security, and digital forensic. He is a member of the IEEE since Mar. 1, 2014.
\end{IEEEbiography}

\bf\vspace{-33pt}
\begin{IEEEbiography}[{\includegraphics[width=1in,height=1.25in,clip,keepaspectratio]{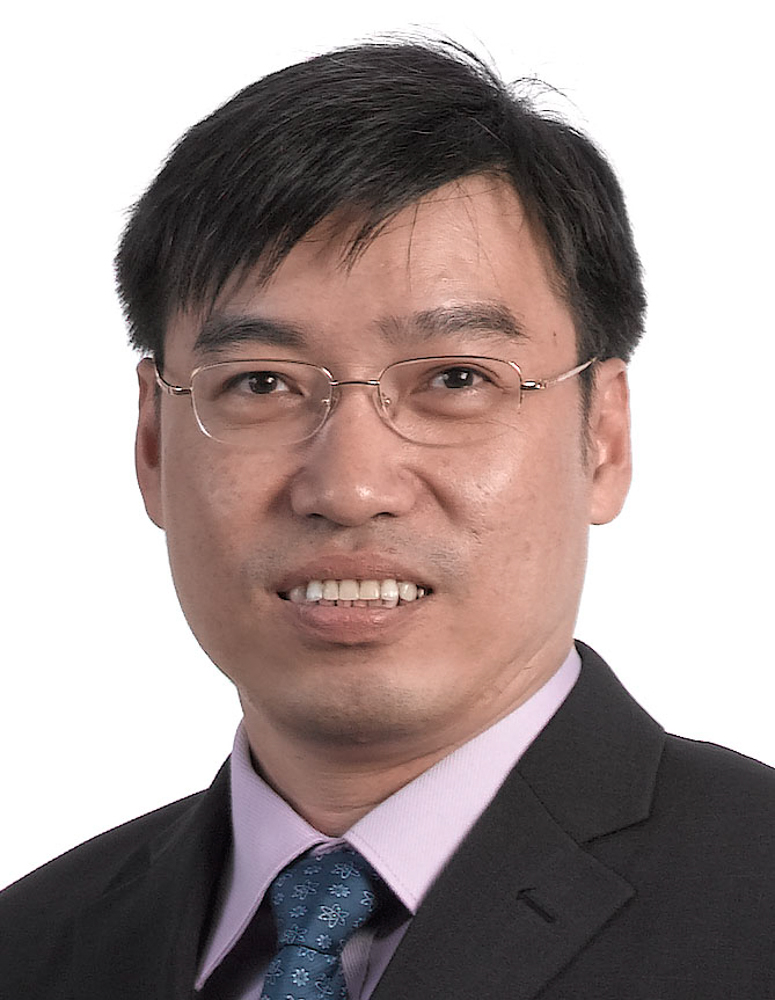}}]{Chip-Hong Chang}(S'92-M'98-SM'03-F'18) received the B.Eng. (Hons.) degree from the National University of Singapore in 1989, and the M. Eng. and Ph.D. degrees from Nanyang Technological University (NTU) of Singapore in 1993 and 1998, respectively. He is a Professor of the School of Electrical and Electronic Engineering (EEE) of NTU. He held joint appointments with the university as Assistant Chair of Alumni from 2008 to 2014, Deputy Director of the Center for High Performance Embedded Systems from 2000 to 2011, and Program Director of the Center for Integrated Circuits and Systems from 2003 to 2009. He has coedited 6 books, published 13 book chapters, more than 120 international journal papers (\textgreater90 are in IEEE) and more than 200 refereed international conference papers (mostly in IEEE), and delivered over 60 keynotes, tutorials and invited seminars. His current research interests include hardware security, AI security, biometric security, trustworthy sensing and hardware accelerators for post-quantum cryptography and edge computational intelligence.

Dr. Chang currently serves as the Senior Area Editor of IEEE Transactions on Information Forensic and Security (TIFS) and IEEE Journal on Emerging and Selected Topics in Circuits and Systems, and Associate Editor of IEEE Transactions on Very Large Scale Integration (VLSI) Systems. He also served as the Associate Editor of TIFS, IEEE Transactions on Computer-Aided Design of Integrated Circuits and Systems, IEEE Access, and IEEE Transactions on Circuits and Systems-I. He is an IEEE Fellow, IET Fellow, AAIA Fellow and the 2018-2019 IEEE Circuits and Systems Society Distinguished Lecturer.\end{IEEEbiography}

\vspace{11pt}

\vfill

\end{document}